\documentclass{article}

\usepackage[preprint]{neurips_2026}

\usepackage[utf8]{inputenc}     
\usepackage[T1]{fontenc}        
\usepackage{microtype}          
\usepackage{nicefrac}           

\usepackage{amsmath, amssymb}   
\usepackage{amsfonts}           

\usepackage[table]{xcolor}      
\definecolor{color3}{RGB}{240,240,250}
\definecolor{citeblue}{RGB}{0,0,128}
\definecolor{softgreen}{RGB}{0,150,0}
\usepackage{url}                
\usepackage{hyperref}           
\hypersetup{
    colorlinks = true,
    citecolor  = citeblue,      
    linkcolor  = red,           
    urlcolor   = blue,          
}

\usepackage{graphicx}           
\usepackage{caption}            
\usepackage{subcaption}         
\usepackage{wrapfig}            
\usepackage{stfloats}           

\usepackage{booktabs}           
\usepackage{multirow}           
\usepackage{arydshln}           
\usepackage{adjustbox}          

\usepackage{algorithm}
\usepackage{algpseudocode}

\title{SOAR: Scale Optimization for Accurate Reconstruction in NVFP4 Quantization}
\vspace{-3mm}
\author{%
  \textbf{Chengzhu Bao}\textsuperscript{*,1} \quad 
  \textbf{Xianglong Yan}\textsuperscript{*,1} \quad 
  \textbf{Zhiteng Li}\textsuperscript{1} \quad 
  \textbf{Guangshuo Qin}\textsuperscript{1} \\
  \textbf{Guanghua Yu}\textsuperscript{2} \quad 
  \textbf{Yulun Zhang}\textsuperscript{$\dagger$,1} \\
  \vspace{-2mm} \\
  \textsuperscript{1}Shanghai Jiao Tong University \quad \textsuperscript{2}Tencent Hunyuan
}

\begin{document}

\maketitle
\renewcommand{\thefootnote}{\fnsymbol{footnote}}
\footnotetext[1]{Equal contribution.} 
\renewcommand{\thefootnote}{\arabic{footnote}}

\makeatletter
{\renewcommand{\thefootnote}{$\dagger$}
\footnotetext{Correspondence to: Yulun Zhang \texttt{<yulun100@gmail.com>}}}
\makeatother

\vspace{-5mm}
\begin{abstract}
NVFP4 has recently emerged as an efficient 4-bit microscaling format for large language models (LLMs), offering superior numerical fidelity with native hardware support. However, existing methods often yield suboptimal performance due to inflexible scale selection and the coupled treatment of quantization and dequantization scales. To address these issues, we propose \underline{S}cale \underline{O}ptimization for \underline{A}ccurate \underline{R}econstruction (SOAR), a novel post-training quantization framework that improves the accuracy of NVFP4 quantization. At its core, SOAR features Closed-form Joint Scale Optimization (CJSO), which jointly optimizes global and block-wise scales via analytical solutions derived from reconstruction error minimization. Furthermore, it incorporates Decoupled Scale Search (DSS). DSS decouples the high-precision quantization scale from its constrained dequantization counterpart, and performs discrete search to mitigate precision loss from scale quantization. Extensive experiments across multiple LLMs show that our method consistently outperforms existing NVFP4 quantization baselines, achieving superior accuracy under the same memory footprint with no additional hardware overhead. The code and models will be available at \url{https://github.com/steven-bao1/SOAR}
\end{abstract}
\setlength{\abovedisplayskip}{2pt}
\setlength{\belowdisplayskip}{2pt}

\vspace{-3.5mm}
\section{Introduction}
\vspace{-2mm}
\begin{wrapfigure}{r}{0.43\textwidth}
\vspace{-14.8mm}
 \centering
\includegraphics[trim=0mm 0mm 0mm 0mm, clip, width=0.43\textwidth]{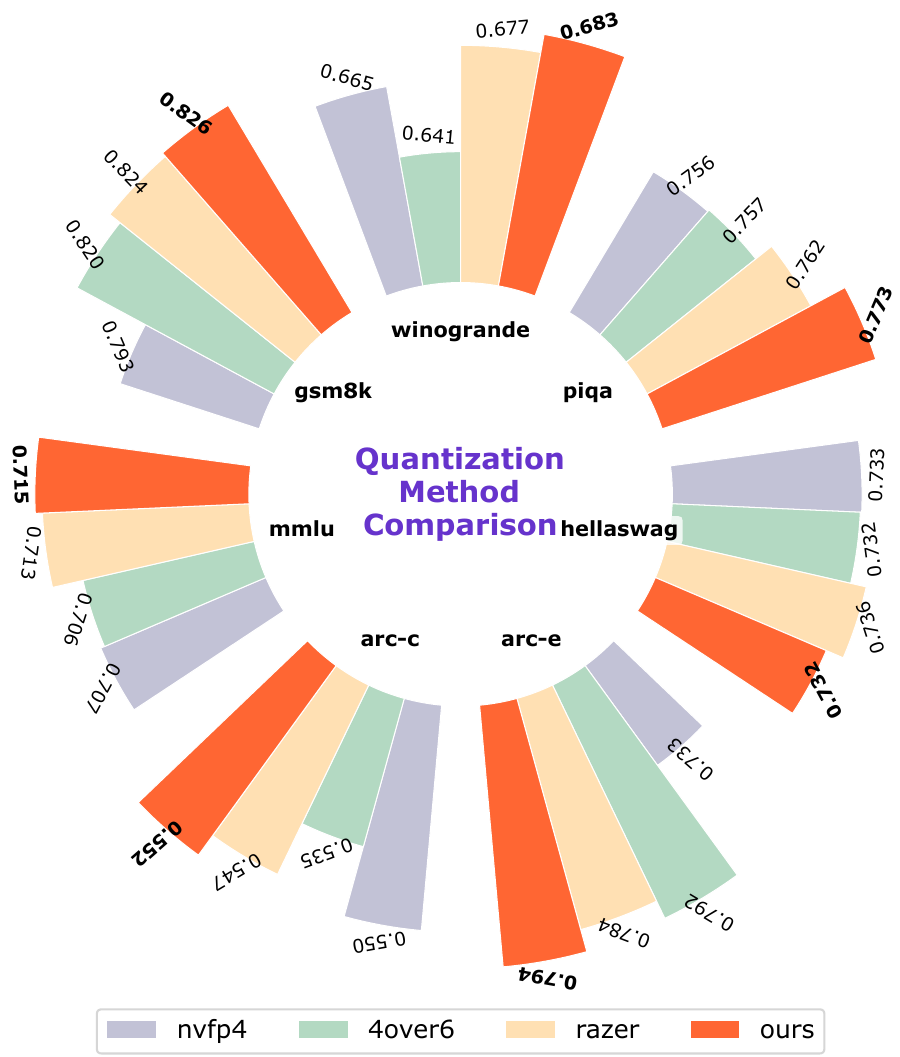}
\vspace{-4.5mm}
\caption{Zero-shot performance of Qwen3-8B under NVFP4 Quantization.}
\label{fig:comparison}
\vspace{-7.5mm}
\end{wrapfigure} 
Large language models (LLMs) have achieved remarkable success in natural language processing (NLP) tasks, exhibiting strong capabilities in both semantic understanding and content generation. Yet, this success largely relies on scaling up model size. Therefore, modern LLMs such as Qwen~\citep{qwen3technicalreport} and LLaMA~\citep{grattafiori_llama_2024} continue to grow to further improve performance. However, the high memory and computational demands of LLMs make deployment difficult, which poses a significant challenge for edge computing and applications that need low latency.

To tackle these deployment challenges, quantization~\citep{dettmers2022gpt3,frantar_gptq_2023} has become a widely adopted solution for efficient LLM inference. By representing parameters with lower bit-widths, quantization significantly reduces the memory footprint and accelerates computation on specialized low-precision hardware. While prior research has predominantly focused on standard integer formats (e.g., INT8 and INT4)~\citep{ashkboos_quarot_2024,frantar_gptq_2023,lin_awq_2024,xiao_smoothquant_2024,li2024arb,yan2025pt,yan2026d2quant}, recent literature has shifted toward low-precision floating-point formats~\citep{liu2023llmfp,chmiel2026fp} which better adapt to LLM tensor distributions. Among these, the microscaling formats--MXFP4~\citep{rouhani2023microscaling} and NVFP4~\citep{abecassis2025pretraining} have gained substantial attention. These formats implement a hierarchical quantization strategy, where elements within a block share a unified scaling factor. Such a floating-point representation better matches the heavy-tailed weight distributions in LLMs, while the fine-grained block-wise scaling further improves local reconstruction accuracy. Moreover, these microscaling formats are natively supported by modern hardware such as NVIDIA Blackwell GPUs~\citep{nvidia2024blackwell}, enabling high-performance LLM deployment.

\begin{figure*}[t]
  \centering
  \vspace{-1mm}
  
  \begin{subfigure}[b]{0.45\textwidth}
    \centering
    \includegraphics[trim=2mm 0mm 2mm 0mm, clip, width=\textwidth]{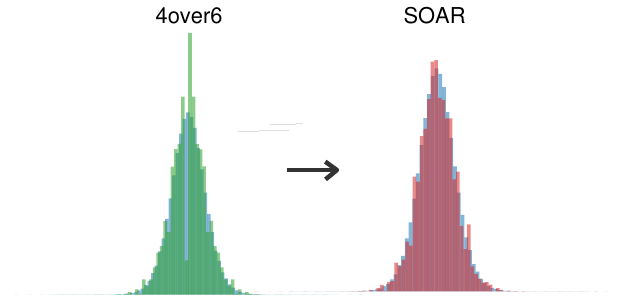}
    \caption{Comparison of weight distributions}
    \label{fig:problem_1}
  \end{subfigure}
  \hfill 
  \begin{subfigure}[b]{0.48\textwidth}
    \centering
    \includegraphics[trim=2mm 0mm 2mm 0mm, clip, width=\textwidth]{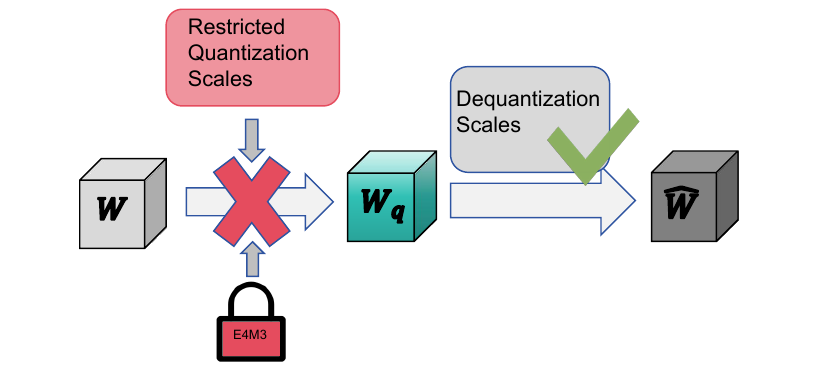}
    \caption{The coupled scaling problem}
    \label{fig:problem_2}
  \end{subfigure}

  \vspace{-1mm}
  \caption{ Motivations for SOAR.
\textbf{(a)} Comparison of weight distributions: Current scaling strategy provides a sub-optimal fit for LLM weights, whereas SOAR adaptively fits the weight distribution. \textbf{(b)} The coupled scaling problem: Hardware constraints on dequantization scales traditionally restrict quantization scales, SOAR resolves this by decoupling the two processes.}
  \label{fig:problem_total}
  \vspace{-9mm}
\end{figure*}

Driven by these advantages, a growing body of research has focused on developing quantization methods tailored for the microscaling regime. 4over6~\citep{cook2025four} and MR-GPTQ~\citep{mr-gptq2026} both determines scaling factors by searching based on MSE, while the latter combines GPTQ~\citep{frantar_gptq_2023} and blockwise Hadamard rotation. Other work~\citep{shao2025block,li2026batquant} also underscores the importance of blockwise rotation and introduces block-wise transformations to mitigate outlier-induced performance collapse. Latest work~\citep{chhugani2026unveiling} has optimized scaling mechanisms by leveraging overflow-aware and macro-block strategies to extend the dynamic range without hardware modifications. Moreover, RaZeR~\citep{chen2025razer} introduces redundant zero remapping, which leverages unused bits in NVFP4 to broaden the numerical coverage, achieving higher accuracy within the same memory footprint. As illustrated in Fig.~\ref{fig:comparison}, these methods have improved NVFP4 quantization performance to some extent.

However, despite the extensive optimizations proposed for 4-bit microscaling formats, several critical bottlenecks remain unaddressed within this regime, as shown in Fig.~\ref{fig:problem_total}. Current methodologies primarily rely on deterministic scaling rules to compute the global and block-wise scaling factors. Even adaptive strategies, such as 4over6 \citep{cook2025four}, are restricted to a simple binary choice between predefined ranges, which lacks the sufficient resolution to capture the complex and irregular weight distributions often observed in large-scale models. Furthermore, existing microscaling quantization methods are constrained by hardware-dependent block-wise scaling, where a single quantized scale is shared between quantization and dequantization, causing scale quantization error to affect both processes. Specifically, while dequantization must follow hardware constraints, the quantization process itself could benefit from a higher-precision scaling representation, since the final representation only consists of an FP4 weight matrix and a quantized dequantization scale.

To address these issues, we propose \underline{S}cale \underline{O}ptimization for \underline{A}ccurate \underline{R}econstruction (SOAR), a principled Post-Training Quantization (PTQ) method that improves NVFP4 quantization. Firstly, we propose Closed-form Joint Scale Optimization (CJSO), an iterative optimization framework that jointly updates the global scale and block-wise scales. Specifically, we derive closed-form updates by differentiating the reconstruction error with respect to the global scale and block-wise scales, enabling joint optimization of both scales. We further propose Decoupled Scale Search (DSS) to mitigate the accuracy loss caused by E4M3 quantization of block-wise scales in NVFP4. In contrast to directly optimizing block scales, DSS decouples the scaling process into a full-precision quantization scale and a hardware-constrained dequantization scale that simulates inference-time E4M3 quantization. Based on this decoupled formulation, we design a discrete search strategy that jointly explores the candidate spaces of both scaling factors, and selects the configuration that minimizes the reconstruction error. Together, these two components form SOAR, integrating CJSO with DSS for NVFP4 quantization. As shown in Fig.~\ref{fig:comparison}, on Qwen3-8B under NVFP4 quantization, SOAR achieves an accuracy of 70.68 over 5 zero-shot tasks (vs. 70.12 for the state of the art (SOTA)). 

Our main contributions are summarized as follows: 
\vspace{-1mm}
\begin{itemize}
\vspace{-1mm}
\item We propose SOAR, a novel NVFP4 PTQ framework that enables accurate reconstruction without incurring additional hardware overhead, addressing the scale optimization limitations of existing microscaling methods.
\vspace{-1mm}
\item We propose Closed-form Joint Scale Optimization (CJSO), which jointly optimizes global and block-wise scales through analytical closed-form updates, improving reconstruction accuracy and quantization performance.
\vspace{-1mm}
\item We propose Decoupled Scale Search (DSS), which decouples quantization and dequantization scales and refines them through discrete search, reducing the accuracy loss caused by E4M3-constrained block-scale quantization.
\vspace{-1mm}
\item Extensive experiments demonstrate that our method consistently outperforms existing NVFP4 quantization baselines across a wide range of LLMs, achieving superior accuracy under the same memory footprint without additional overhead. 
\end{itemize}

\vspace{-2mm}
\section{Related Works}
\vspace{-2mm}
\subsection{Post-training Quantization for LLMs}
\vspace{-2mm}
Post-training quantization (PTQ) has become one of the most effective approach for efficient LLM deployment due to its ability to reduce memory and computation cost without retraining. Current PTQ strategies can be categorized into mixed-precision, compensation-based, and transformation-based methods. \textbf{Mixed-precision methods} adaptively allocate varying bit-widths based on quantization sensitivity.  Quik~\citep{QUIK} and subsequent works~\citep{zhao2024atom,saxena2025resq} keep the majority of parameters in INT4 while preserving a subset of outlier channels in INT8 or FP16. Slim-LLM~\citep{huang_slim-llm_2025} and other works~\citep{2025skim,kim_squeezellm_2024} adopt finer-grained bit allocation strategies with 1-bit granularity, assigning more bits to more important weights. \textbf{Compensation-based methods} such as GPTQ~\citep{frantar_gptq_2023} reduce reconstruction error by employing Hessian-based optimization to adjust full-presision weights. Building on this, GPTAQ~\citep{li_gptaq_2025} and QEP~\citep{arai_quantization_2026} account for input errors, while BOA~\citep{kim_boa_nodate} optimizes Hessian estimation, further boosting accuracy. \textbf{Transformation-based methods} apply equivalent transformations to suppress outliers and smooth tensor distributions prior to quantization. Channel-wise scaling methods such as AWQ~\citep{lin_awq_2024} and SmoothQuant~\citep{xiao_smoothquant_2024} apply per-channel rescaling to mitigate quantization difficulty in low-bit regimes. Methods like Quarot~\citep{ashkboos_quarot_2024} and a series of works~\citep{lin2024duquant,shao2026dartquant,liang2026paroquant,hu2025ostquant,sun2024flatquant,ma2024affinequant} employ rotation transforms to flatten activation distributions, improving reconstruction accuracy. However, these methods are designed for integer quantization and often underperform when applied to the microscaling format, due to its distinct numerical properties and hardware-constrained scaling.

\begin{figure*}[t]
  \centering
  \vspace{-1mm}
   \includegraphics[width=1.0\textwidth]{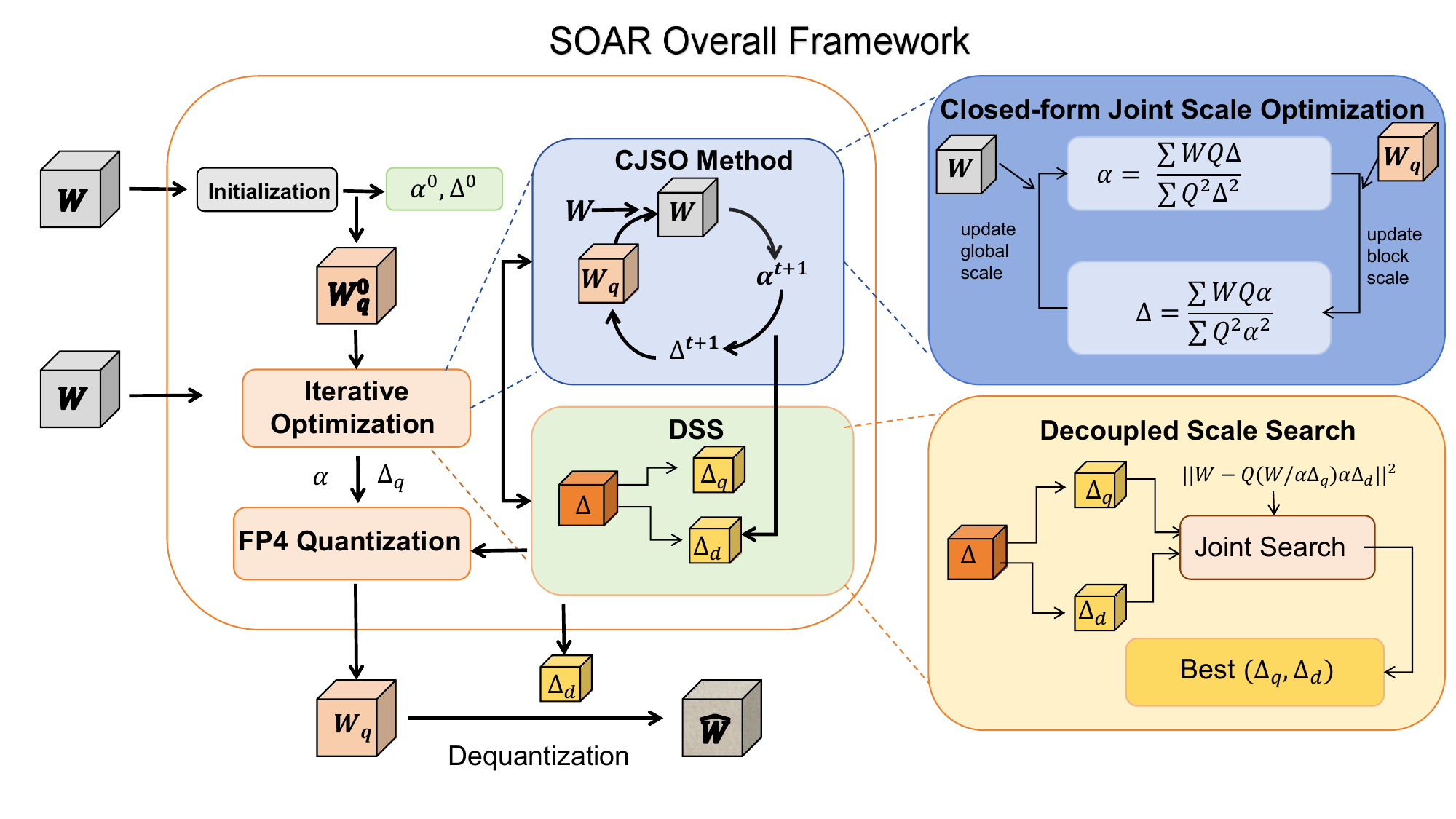}
   \vspace{-5mm}
    \caption{Overview of SOAR. The left panel shows the iterative SOAR framework for scale refinement. The top-right illustrates CJSO, which provides updates for joint global and block-wise scale optimization. The bottom-right depicts DSS, which decouples quantization and dequantization scales to mitigate precision loss via joint discrete search.}
   \label{fig:overview}
   \vspace{-4mm}
\end{figure*}

\vspace{-1mm}
\subsection{Microscaling Floating Point Formats}
\vspace{-1mm}
Recently, Microscaling FP4 Formats--MXFP4 and NVFP4 have emerged as efficient quantization regimes, where parameters are quantized with shared block-wise scaling factors. With native hardware support, they have become promising choices for efficient low-bit LLM deployment, and several works have introduced optimizations tailored for the formats. 4over6~\citep{cook2025four} adaptively chooses the scaling range of NVFP4 blocks between 6 and 4 to reduce quantization error. MR-GPTQ~\citep{mr-gptq2026} further refines scaling factors through reconstruction-aware search, and combines GPTQ~\citep{frantar_gptq_2023} with block-wise Hadamard rotation to improve FP4 quantization. Other work~\citep{shao2025block,li2026batquant} also show that global rotation may even degrade model performance, while block-wise transformations are more effective in mitigating outlier-induced accuracy loss. Latest work~\citep{chhugani2026unveiling} introduces Overflow-Aware Scaling and Macro Block Scaling to improve MXFP4 fidelity by enhancing dynamic range without hardware modifications. Moreover, ARCQuant~\citep{meng2026arcquant} enhances NVFP4 performance by augmenting activations with quantized residual channels. RaZeR~\citep{chen2025razer} enhances NVFP4 by remapping redundant zeros to additional quantization values, improving accuracy under the same memory footprint. However, current methods use simple and fixed rules to determine block-wise scales, but recent research~\citep{zhang2026benchmarking} has identified scaling factor inaccuracies as an error source. Moreover, hardware-constrained block scales affect both quantization and dequantization, introducing additional error. To address this, our method analytically determines scaling factors and decouples the two processes, improving accuracy under NVFP4 quantization.

\vspace{-1mm}
\section{Method}
\vspace{-2mm}
In this section, we introduce our method SOAR, as illustrated in Fig.~\ref{fig:overview}. We first review preliminaries of NVFP4 quantization in Sec~\ref{3.1}. Then, Sec~\ref{3.2} presents the Closed-form Joint Scale Optimization (CJSO) for joint global and block-wise scale optimization, followed by Sec~\ref{3.3}, which introduces the Decoupled Scale Search (DSS) to address hardware-constrained block-wise scaling. Finally, Sec~\ref{3.4} combines CJSO and DSS into our overall framework.

\vspace{-2mm}
\subsection{Preliminary}
\label{3.1}
\vspace{-2mm}
NVFP4 is a hierarchical block-wise low-precision floating-point format that represents weights using FP4 (E2M1) elements with a two-level scaling mechanism, consisting of a tensor-wide FP32 global scale and FP8 (E4M3) block-wise scales applied every 16 elements. For a high-precision tensor $X$, the quantization process can be formulated as follows: 
\begin{equation}
\alpha = \frac{\max(|X|)}{M_{\text{FP4}} \cdot M_{\text{FP8}}}, \quad
\Delta_i = \mathcal{Q}_{\text{E4M3}}\!\left(
\frac{\max\left(|X_i|\right)}{\alpha \cdot M_{\text{FP4}}}
\right),
\end{equation}
\begin{equation}
\bar{X} =
\begin{cases}
\frac{1}{2} \left\lceil \frac{2X}{\alpha \Delta} \right\rfloor, & \left| \frac{X}{\alpha \Delta} \right| < 2 \\
\left\lceil \frac{X}{\alpha \Delta} \right\rfloor, & 2 \le \left| \frac{X}{\alpha \Delta} \right| < 4 \\
2 \left\lceil \frac{X}{2\alpha \Delta} \right\rfloor, & 4 \le \left| \frac{X}{\alpha \Delta} \right| \le 6.
\end{cases}
\end{equation}
$M_{\text{FP4}}$ and $M_{\text{FP8}}$ denote the maximum representable values of FP4 (E2M1) and FP8 (E4M3), 6 and 448 respectively, and $\mathcal{Q}_{\text{E4M3}}$ denotes the FP8 (E4M3) quantization function, and $\lceil \cdot \rfloor$ denotes the rounding-to-nearest function. The corresponding dequantization process is given by:
\begin{equation}
\hat{X} = \bar{X} \cdot (\alpha \Delta),
\end{equation}
where $\Delta$ denotes the block-wise scaling vector applied to each 16-element group.
\vspace{-1mm}
\subsection{Closed-form Joint Scale Optimization}
\label{3.2}
\vspace{-2mm}
\textbf{Motivation.} Existing NVFP4 quantization methods typically determine the global scale and block-wise scales using simple heuristic formulas, such as max-based scaling, or limited discrete search over a limited candidate set, which often lead to suboptimal results. In particular, for the FP32 global scale, restricting its optimization to fixed rules or a coarse discrete space makes it difficult to accurately capture the complex weight distribution of large language models. This motivates the need for a more principled, analytical approach that treats scale determination as a joint optimization problem.

To address this, we propose \textbf{Closed-form Joint Scale Optimization (CJSO)}, which jointly optimizes the global scale and block-wise scales by directly minimizing the reconstruction error between the original weights $W$ and the dequantized weights $\hat{W}$. Specifically, we formulate the optimization objective as:
\begin{equation}
\min_{\alpha, \{\Delta_i\}} \mathcal{L} = \sum_{i} \left\| W_i - Q_i \cdot (\alpha \Delta_i) \right\|_2^2,
\end{equation}
where $Q_i = \mathcal{Q}_{\text{FP4}}(W_i / (\alpha \Delta_i))$ represents the FP4 quantization. 

Directly optimizing the reconstruction objective is challenging since $Q_i$ is discrete and depends on both $\alpha$ and $\Delta_i$. However, under a fixed assignment of $Q_i$, the reconstruction objective becomes a quadratic function with respect to the scaling factors. This allows us to derive closed-form updates by satisfying the first-order optimality conditions $\frac{\partial \mathcal{L}}{\partial \alpha} = 0$ and $\frac{\partial \mathcal{L}}{\partial \Delta_i} = 0$. 

\textbf{Global Scale Optimization.} Given the block-wise scales $\Delta_i$, the optimal tensor-wide scale $\alpha^*$ is obtained as:
    \begin{equation}
    \alpha^* = \frac{\sum_{i=1}^N \sum_{j \in \text{block}_i} W_{ij} \cdot Q_{ij} \cdot \Delta_i}{\sum_{i=1}^N \sum_{j \in \text{block}_i} Q_{ij}^2 \cdot \Delta_i^2}.
    \end{equation}
\textbf{Block-wise Scale Optimization.} Conversely, With $\alpha$ fixed, each block-wise scale $\Delta_i^*$ can be independently optimized to fit its local distribution:
    \begin{equation}
    \Delta_i^* = \frac{\sum_{j \in \text{block}_i} W_{ij} \cdot Q_{ij} \cdot \alpha}{\sum_{j \in \text{block}_i} Q_{ij}^2 \cdot \alpha^2},
    \end{equation}
 $\Delta_i^*$ is subsequently projected onto the nearest value in the FP8 (E4M3) format.

\textbf{Refinement Scheme.} The process of determining the optimal scaling factors and the FP4 assignments $Q$ forms a coordinate-wise optimization framework. We initialize $\alpha$ and $\Delta_i$ using the standard max-based rules in NVFP4. Then, the global scale $\alpha$, block-wise scales $\Delta_i$, and the quantized matrix $Q$ are updated iteratively: after each scale update, $Q$ is recomputed by applying FP4 quantization under the updated scaling factors.

\subsection{Decoupled Scale Search}
\label{3.3}
\textbf{Motivation.} In NVFP4, the block-wise scaling factor is constrained by the limited precision of FP8 (E4M3) representation in hardware, which introduces quantization error itself. Moreover, this error propagates to both quantization and dequantization stages, directly affecting reconstruction accuracy. However,existing methods typically use a unified block-wise scaling factor that is shared between quantization and dequantization, without explicitly distinguishing the two processes. This entanglement prevents independent optimization of the two processes under low-precision constraints, limiting the expressiveness of scale design.

\textbf{Decoupled Scale Formulation.} The NVFP4 format is characterized by a hardware-constrained storage footprint, consisting of a tensor-wise FP32 global scale $\alpha$, a block-wise FP8 (E4M3) dequantization scale $\Delta_i^d$, and a 4-bit (E2M1) weight matrix. Notably, while $\Delta_i^d$ must comply with FP8 precision for hardware compatibility, the scaling used to determine FP4 assignments during quantization is not subject to this constraint, revealing an inherent asymmetry between the quantization and dequantization processes.

Based on this insight, we introduce a decoupled scaling formulation, separating the role of block-wise scaling into a high-precision quantization scale $\Delta_i^q$ and a hardware-constrained dequantization scale $\Delta_i^d$. The former controls the FP4 assignment by influencing the rounding of weights, while the latter is responsible for reconstruction.

Formally, we reformulate NVFP4 quantization as:
\begin{equation}
\min_{\Delta_i^q \in \mathbb{R}, \Delta_i^d \in \text{FP8}} 
\mathcal{L}
=
\sum_i
\left\|
W_i -
\mathcal{Q}_{\text{FP4}}\!\left(\frac{W_i}{\alpha \Delta_i^q}\right)
\cdot (\alpha \Delta_i^d)
\right\|_2^2.
\end{equation}

In this formulation, $\Delta_i^q$ determines how weights are mapped to FP4 levels during quantization, while $\Delta_i^d$ is the actual stored scaling factor under hardware constraints. This decoupling allows the quantization process to be optimized in a high-precision space without being constrained by the low-precision scale used for reconstruction.

\textbf{Decoupled Joint Search.}
Based on the decoupled formulation, we optimize the quantization-side scale $\Delta_i^q$ and the dequantization-side scale $\Delta_i^d$ jointly for each block. Since the two scales jointly affect the final reconstruction error, optimizing either one independently often leads to suboptimal FP4 assignments. We therefore formulate the problem as a local joint search over the scale pair.

\textbf{(i) Initialization:} We initialize the global scale $\alpha$ and block-wise dequantization scale $\Delta_i^d$ using the max-based rule in Sec. 3.1, with the initial quantization scale set to $\Delta_i^q = \Delta_i^d$.

\textbf{(ii) Joint Scale Refinement:} Starting from this initialization, we perform a block-wise local joint search to refine the scale pair $(\Delta_i^q, \Delta_i^d)$. For the dequantization scale, candidate values are restricted to nearby FP8-representable values, since it must remain hardware-compliant for final storage. On the other hand, the quantization-side scale is searched in high precision using continuous multiplicative perturbations around the initialization, allowing more flexible control of FP4 assignments. For each candidate pair $(\Delta_i^q, \Delta_i^d)$, we evaluate the reconstruction objective:
\begin{equation}
\mathcal{L}_i =
\left\|
W_i -
\mathcal{Q}_{\text{FP4}}\!\left(\frac{W_i}{\alpha \Delta_i^q}\right)
\cdot (\alpha \Delta_i^d)
\right\|_2^2,
\end{equation}
and select the pair that minimizes the block-wise reconstruction error. In this way, $\Delta_i^q$ determines a more suitable FP4 mapping, while $\Delta_i^d$ preserves hardware compatibility for dequantization.

\begin{figure*}[t]
  \centering
  \vspace{-1mm}
   \includegraphics[width=1.0\textwidth]{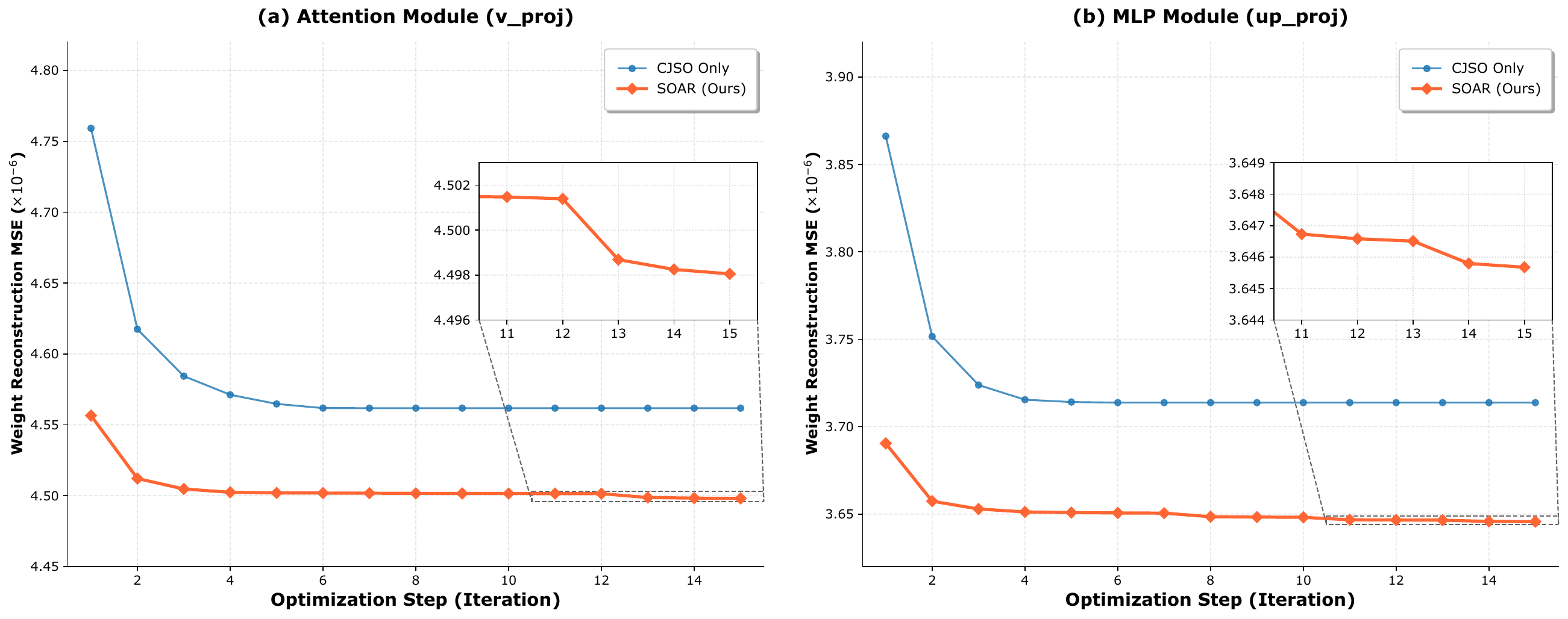}
   \vspace{-6mm}
    \caption{Convergence of reconstruction MSE. (a) and (b) show the optimization progress on Attention and MLP modules, respectively. SOAR consistently reaches lower error floors than using CJSO strategy only.}
   \label{fig:mse_error}
   \vspace{-6mm}
\end{figure*}
\vspace{2mm}

\vspace{-2mm}
\subsection{SOAR Framework}
\label{3.4}
\vspace{-2mm}

As shown in Algorithm in Appendix~\ref{alg}, we combine CJSO and DSS into a unified iterative optimization framework tailored for NVFP4 quantization, where each iteration consists of two stages: closed-form analytical scale update and decoupled local search.

\textbf{Closed-form Update.} Starting from the initialized scales, each iteration consists of two stages. First, we update the tensor-wise global scale $\alpha$ and the block-wise scale $\Delta_i$ using one-step closed-form optimization under the current FP4 assignments. This step quickly moves the optimization toward a low-error region and provides an accurate prior for subsequent search.

\textbf{Decoupled Search.} Next, based on the updated block-wise scale, we perform decoupled scale refinement using DSS. For each block, the dequantization scale $\Delta_i^d$ is restricted to the nearest FP8-representable neighbors around the current solution, avoiding excessive search overhead while preserving hardware compatibility. The quantization scale $\Delta_i^q$ is then searched in high precision within this local region to obtain a better FP4 assignment. The pair that minimizes the reconstruction error is selected for the next iteration.

By alternating analytical optimization and local decoupled search, the framework combines the efficiency of closed-form scale estimation with the flexibility of fine-grained reconstruction-aware refinement. In particular, CJSO provides a strong prior for DSS, while DSS compensates for the precision loss introduced by hardware-constrained FP8 scales, leading to more accurate NVFP4 quantization. The convergence of this iterative process is illustrated in Fig.~\ref{fig:mse_error}, where SOAR consistently achieves lower MSE across iterations compared to the CJSO-only baseline.

\begin{table*}[t]
\setlength{\tabcolsep}{5pt}
\small
\centering
\caption{Main results on NVFP4 quantization.We evaluate models under the W4A4 quantization setting and report zero-shot accuracy ($\uparrow$) on five commonsense tasks, along with their average (Avg.). Best results are in \textbf{bold}.}
\label{tab:main_nvfp4}
\vspace{0.5mm}
\begin{adjustbox}{max width=\linewidth}
\begin{tabular}{c|l|cccccc}
\hline
\rowcolor{color3}
\textbf{Model} & \textbf{Method} &
\textbf{WinoGrande} & \textbf{PIQA} & \textbf{HellaSwag} & \textbf{Arc-E} & \textbf{Arc-C} & \textbf{Avg.} \\
\hline

\multirow{5}{*}{\shortstack{LLaMA3.1\\8B-Instruct}}
& FP16    & 74.43 & 81.01 & 79.30 & 82.87 & 54.78 & 74.48 \\
\cdashline{2-8}
& NVFP4   & 71.82 & 78.84 & 77.98 & 78.70 & 52.22 & 71.91 \\
& 4over6    & 71.03 & 80.25 & 77.96 & 78.91 & 52.30 & 72.09 \\
& RaZeR   & 71.51 & \textbf{80.36} & \textbf{78.13} & \textbf{79.76} & 52.90 & 72.53 \\
& SOAR   & \textbf{72.30} & 79.71 & 77.86 & 78.87 & \textbf{54.10} & \textbf{72.57} \\
\hline

\multirow{5}{*}{\shortstack{LLaMA3.2\\1B-Instruct}}
& FP16   2 & 61.64 & 74.92 & 61.59 & 63.72 & 37.54 & 59.88 \\
\cdashline{2-8}
& NVFP4   & 59.19 & 71.22 & 57.97 & 59.64 & 35.07 & 56.62 \\
& 4over6   & \textbf{60.30} & 71.38 & 57.78 & 60.02 & 36.43 & 57.18 \\
& RaZeR   & 59.51 & \textbf{72.09} & \textbf{59.00} & 60.94 & 36.01 & 57.51 \\
& SOAR    & 59.91 & 71.22 & 58.34 & \textbf{61.24} & \textbf{38.23} & \textbf{57.79} \\
\hline

\multirow{5}{*}{\shortstack{LLaMA3.2\\3B-Instruct}}
& FP16    & 66.83 & 76.66 & 71.55 & 71.09 & 46.25 & 66.83 \\
\cdashline{2-8}
& NVFP4   & 65.11 & 75.63 & 70.38 & 69.36 & 44.62 & 65.02 \\
& 4over6     & 65.43 & 75.63 & 70.03 & 69.28 & 43.34 & 64.74 \\
& RaZeR   & 67.01 & 76.17 & \textbf{70.79} & 67.85 & 44.11 & 65.19 \\
& SOAR    & \textbf{67.88} & \textbf{76.17} & 70.70 & \textbf{70.03} & \textbf{45.22} & \textbf{66.00} \\
\hline

\multirow{5}{*}{\shortstack{Qwen3\\8B}}
& FP16    & 67.64 & 77.58 & 74.86 & 80.89 & 56.66 & 71.53 \\
\cdashline{2-8}
& NVFP4   & 66.54 & 75.63 & 73.27 & 73.27 & 55.03 & 68.75 \\
& 4over6     & 64.09 & 75.68 & 73.24 & 79.25 & 53.50 & 69.15 \\
& RaZeR   & 67.72 & 76.22 & \textbf{73.58} & 78.41 & 54.69 & 70.12 \\
& SOAR    & \textbf{68.35} & \textbf{77.26} & 73.19 & \textbf{79.42} & \textbf{55.20} & \textbf{70.68} \\
\hline

\multirow{5}{*}{\shortstack{Qwen3\\4B}}
& FP16   & 65.9 & 75.08 & 68.46 & 78.45 & 54.01 & 68.38 \\
\cdashline{2-8}
& NVFP4   & 61.01 & 73.07 & 65.94 & 74.49 & 50.94 & 65.09 \\
& 4over6     & 60.3 & 71.65 & 66.11 & 73.06 & 48.04 & 63.83 \\
& RaZeR   & \textbf{63.77} & 73.61 & 66.54 & \textbf{74.79} & 50.26 & 65.79 \\
& SOAR   & 63.14 & \textbf{74.32} & \textbf{66.68} & 74.24 & \textbf{50.94} & \textbf{65.86} \\
\hline

\end{tabular}
\end{adjustbox}
\end{table*}

\vspace{-2mm}
\section{Experiments}
\vspace{-2mm}

\subsection{Experimental Setup}
\vspace{-1mm}

\noindent \textbf{Implementation Details.}
All experiments are conducted using PyTorch~\citep{paszke2019pytorch} and the HuggingFace Transformers library~\citep{wolf-etal-2020-transformers} on NVIDIA A800-80GB GPUs. For experiments involving calibration-based methods such as GPTQ, we use 128 samples from the WikiText-2 dataset~\citep{merity_pointer_2016} with a sequence length of 2048 as the calibration set. We implement 15 iterations for SOAR to ensure the convergence of optimization, with early stopping triggered if the relative MSE improvement falls below $10^{-3}$. More details are in Appendix~\ref{experiment_settings}.

\vspace{-1mm}
\noindent \textbf{Baselines.}
We compare SOAR with 4over6\citep{cook2025four}, and RaZeR~\citep{chen2025razer}, two representative methods in NVFP4 quantization. Since SOAR itself is calibration data-free and does not rely on reconstruction-aware calibration-data, we further combine it with GPTQ~\citep{frantar_gptq_2023}, and compare against the original GPTQ framework.

\vspace{-1mm}
\noindent \textbf{Models and Evaluation.}
We evaluate SOAR on a range of pretrained LLMs, including LLaMA-3.1 (8B)-Instruct~\citep{grattafiori_llama_2024}, LLaMA-3.2 (1B/3B)-Instruct, and Qwen3 (4B/8B)~\citep{qwen3technicalreport}. We assess quantized models using both language modeling and downstream benchmarks. We report zero-shot accuracy on WinoGrande~\citep{sakaguchi_winogrande_2019}, PIQA~\citep{bisk_piqa_2019}, HellaSwag~\citep{zellers_hellaswag_2019}, and ARC-Easy/Challenge~\citep{clark_think_2018}, and evaluate on MMLU~\citep{hendrycks_measuring_2021} and GSM8K~\citep{cobbe2021training}, two comprehensive reasoning tasks spanning multi-domain knowledge and mathematical logic.

\vspace{-2mm}
\subsection{Main Results} 
\vspace{-2mm}
\noindent \textbf{Zero-Shot Accuracy Evaluation.} We evaluate SOAR on five representative zero-shot reasoning benchmarks. As shown in Tab.~\ref{tab:main_nvfp4}, SOAR consistently outperforms existing NVFP4 baselines across most tasks and model sizes. On the Qwen series, for example, SOAR raises the average accuracy of Qwen3-8B from 68.75 (NVFP4) to 70.68, achieving a significant +1.93 gain and surpassing the RaZeR baseline. Similar improvements are observed in the LLaMA series; on LLaMA3.2-3B-Instruct, our method boosts the average accuracy from 65.19 (RaZeR) to 66.00, achieving a +0.81 gain and narrowing the gap to the FP16 baseline. These consistent gains across diverse architectures clearly demonstrate the robust ability of SOAR to preserve the fundamental reasoning and commonsense capabilities of LLMs within the NVFP4 format.

\begin{wraptable}{r}{0.5\textwidth}
\vspace{-4mm}
\captionsetup{width=0.5\textwidth}
\caption{Results on Reasoning tasks. Best results are in \textbf{bold}.}
\label{tab:mmlu_gsm8k}
\centering
\vspace{-2mm}
\resizebox{0.48\textwidth}{!}{%
\begin{tabular}{c|l|cc|c}
\hline
\rowcolor{color3}
\textbf{Model} & \textbf{Method} & \textbf{MMLU} & \textbf{GSM8K} & \textbf{Avg.} \\
\hline
\multirow{5}{*}{\shortstack{LLaMA-3.2\\1B-Instruct}}
& FP16   & 48.25 & 45.26 & 46.76 \\
\cdashline{2-5}
& NVFP4  & 43.28 & 31.77 & 37.53 \\
& 4over6 & 43.98 & 30.93 & 37.46 \\
& RaZeR  & 43.74 & 32.22 & 37.98 \\
& SOAR   & \textbf{44.05} & \textbf{32.83} & \textbf{38.44} \\
\hline
\multirow{5}{*}{\shortstack{LLaMA-3.2\\3B-Instruct}}
& FP16   & 62.30 & 77.10 & 69.70 \\
\cdashline{2-5}
& NVFP4  & 59.07 & 71.19 & 65.13 \\
& 4over6 & 58.43 & 71.80 & 65.12 \\
& RaZeR  & \textbf{59.29} & 72.93 & 66.11 \\
& SOAR   & 59.23 & \textbf{73.62} & \textbf{66.43} \\
\hline
\multirow{5}{*}{\shortstack{Qwen3\\8B}}
& FP16   & 72.93 & 86.05 & 79.49 \\
\cdashline{2-5}
& NVFP4  & 70.70 & 79.30 & 75.00 \\
& 4over6 & 70.61 & 81.96 & 76.29 \\
& RaZeR  & 71.31 & 82.41 & 76.86 \\
& SOAR   & \textbf{71.47} & \textbf{82.56} & \textbf{77.02} \\
\hline
\end{tabular}%
}
\vspace{-8mm}
\end{wraptable}

\noindent \textbf{Reasoning Tasks Evaluation.} To further assess the reasoning and knowledge retention of quantized models, we evaluate SOAR on the MMLU and GSM8K benchmarks, with results summarized in and Tab.~\ref{tab:mmlu_gsm8k}. SOAR improves accuracy across nearly all model sizes. On Qwen3-8B, it achieves a gain of 0.77 points on MMLU and 3.26 points on GSM8K over the standard NVFP4 baseline. On LLaMA-3.2-3B-Instruct, the improvement on GSM8K reaches 0.69 points over RaZeR, significantly narrowing the gap to the FP16 baseline. SOAR effectively mitigates knowledge degradation from NVFP4 quantization, yielding stronger factual and reasoning performance across general-purpose tasks without task-specific adaptation.

\noindent \textbf{Integration with Calibration-Based Methods.} Although SOAR is a calibration-data-free approach, it remains highly compatible with calibration-data-based quantization methods. To demonstrate this, we integrate it with GPTQ to examine its compatibility with calibration-based PTQ frameworks. As summarized in Tab.~\ref{tab:gptq_integration}, SOAR consistently enhances the performance of GPTQ-quantized models across various architectures. For LLaMA-3.1-8B-Instruct, our method elevates the average zero-shot accuracy from 72.95\% to 73.18\%, while maintaining virtually identical perplexity on WikiText2 and C4. Similar additive gains are observed in the LLaMA-3.2 series, where SOAR improves the average accuracy of the 3B model to 65.82\%. The result show that despite being calibration-data-free, SOAR exhibits strong synergy with calibration-data-dependent methods like GPTQ to further push the performance boundaries of NVFP4.
\begin{table*}[h]
\vspace{-2mm}
\setlength{\tabcolsep}{5pt}
\small
\centering
\caption{Integration with GPTQ. We evaluate the compatibility of SOAR with GPTQ. We report perplexity ($\downarrow$) on WikiText2 and C4, and zero-shot accuracy ($\uparrow$) on five commonsense reasoning tasks with their average (Avg.). Best results are in \textbf{bold}.}
\label{tab:gptq_integration}
\vspace{-2mm}

\begin{adjustbox}{max width=\linewidth}
\begin{tabular}{c|l|cc|cccccc}
\hline
\rowcolor{color3}
\textbf{Model} & \textbf{Method} &
\textbf{Wiki2} & \textbf{C4} &
\textbf{Wino.} & \textbf{PIQA} & \textbf{Hella.} & \textbf{Arc-E} & \textbf{Arc-C} & \textbf{Avg.} \\
\hline

\multirow{2}{*}{\shortstack{LLaMA3.1\\8B-Instruct}}
& GPTQ      & 6.57 & 9.40 & 71.59 & 79.82 & \textbf{78.65} & \textbf{80.43} & \textbf{54.27} & 72.95 \\
& Ours+GPTQ & 6.57 & 9.39 & \textbf{73.40} & \textbf{79.92} & 78.43 & 80.30 & 53.84 & \textbf{73.18} \\
\hline

\multirow{2}{*}{\shortstack{LLaMA3.2\\1B-Instruct}}
& GPTQ      & 14.17 & 19.56 & 60.93 & 73.50 & \textbf{60.10} & \textbf{62.37} & \textbf{37.46} & 58.87 \\
& Ours+GPTQ & 14.21 & 19.60 & \textbf{61.88} & \textbf{73.83} & 59.81 & 62.29 & 36.95 & \textbf{58.95} \\
\hline

\multirow{2}{*}{\shortstack{LLaMA3.2\\3B-Instruct}}
& GPTQ      & 11.57 & 14.97 & 66.54 & \textbf{76.50} & 70.63 & 67.72 & \textbf{45.73} & 65.42 \\
& Ours+GPTQ & 11.61 & 14.98 & \textbf{66.77} & 76.39 & \textbf{70.85} & \textbf{69.95} & 45.14 & \textbf{65.82} \\
\hline

\end{tabular}
\end{adjustbox}
\vspace{-2mm}
\end{table*}

\vspace{-2mm}
\subsection{Ablation Study}
\vspace{-2mm}

\noindent \textbf{Effect of CJSO and DSS.} Table~\ref{tab:ablation_cjso_dss} presents a component-wise ablation of SOAR on LLaMA3.2-3B-Instruct under NVFP4 quantization. We evaluate two perplexity metrics (WikiText2 and C4) and the average zero-shot accuracy across five reasoning tasks. Introducing CJSO yields consistent improvements over the NVFP4 baseline, particularly in downstream accuracy (+0.62). DSS further enhances both reconstruction quality and reasoning performance by refining block-wise scales in a decoupled manner. When combined, SOAR achieves the best overall results across all metrics, demonstrating the complementary roles of closed-form optimization and local scale refinement.

\begin{table}[h]
\vspace{-4mm}
\setlength{\tabcolsep}{3.5pt}
\small
\centering
\caption{
Effect of CJSO and DSS. Component-wise breakdown analysis of CJSO and DSS on LLaMA-3.2-3B-Instruct. Best results are in \textbf{bold}.
}
\label{tab:ablation_cjso_dss}

\begin{tabular}{c|l|cc|ccccc|c}
\hline
\rowcolor{color3}
\textbf{Model} & \textbf{Method} &
\textbf{Wiki2} & \textbf{C4} &
\textbf{Wino.} & \textbf{PIQA} & \textbf{Hella.} & \textbf{Arc-E} & \textbf{Arc-C} &
\textbf{Avg.} \\
\hline

\multirow{4}{*}{LLaMA-3.2-3B-Instruct}
& NVFP4  & 11.98 & 15.53 & 65.11 & 75.63 & 70.38 & 69.36 & 44.62 & 65.02 \\
\cdashline{2-10}
& +CJSO  & 12.04 & 15.53 & 67.56 & 75.52 & 70.24 & 69.53 & 45.31 & 65.64 \\
& +DSS   & 11.96 & 15.45 & 66.85 & 76.33 & 70.33 & 68.86 & 45.14 & 65.50 \\
& SOAR   & \textbf{11.88} & \textbf{15.44} & \textbf{67.88} & \textbf{76.17} & \textbf{70.70} & \textbf{70.03} & \textbf{45.22} & \textbf{66.00} \\
\hline

\end{tabular}
\vspace{-3mm}
\end{table}

\noindent \textbf{DSS on MXFP4.} To evaluate the effectiveness of Decoupled Scale Search (DSS), we extend its application to MXFP4, another prominent microscaling regime. Table~\ref{tab:ablation_mxfp4} presents the performance comparison between the standard MXFP4 baseline and its DSS-enhanced version. As expected, decoupling the quantization and dequantization scales consistently yields superior zero-shot performance across multiple model families. On LLaMA-3.1-8B-Instruct, the introduction of DSS achieves a steady improvement of +1.02 points in average accuracy, increasing it from 53.75 to 54.77. Similar gains are observed on LLaMA3.2-3B-Instruct and Qwen3-4B, with average accuracy increases of +0.86 and +0.47, respectively. The results confirm that scale quantization is an important limitation in current microscaling formats, and our DSS provides an effective, generalized solution to mitigate this error while remaining fully compliant with hardware-native constraints.

\vspace{-2mm}
\subsection{Memory Analysis} 
\vspace{-2mm}
\begin{wrapfigure}[12]{r}{0.48\textwidth}
\vspace{-8mm}
 \centering
\includegraphics[trim=3 0 0 0, clip, width=0.48\textwidth]{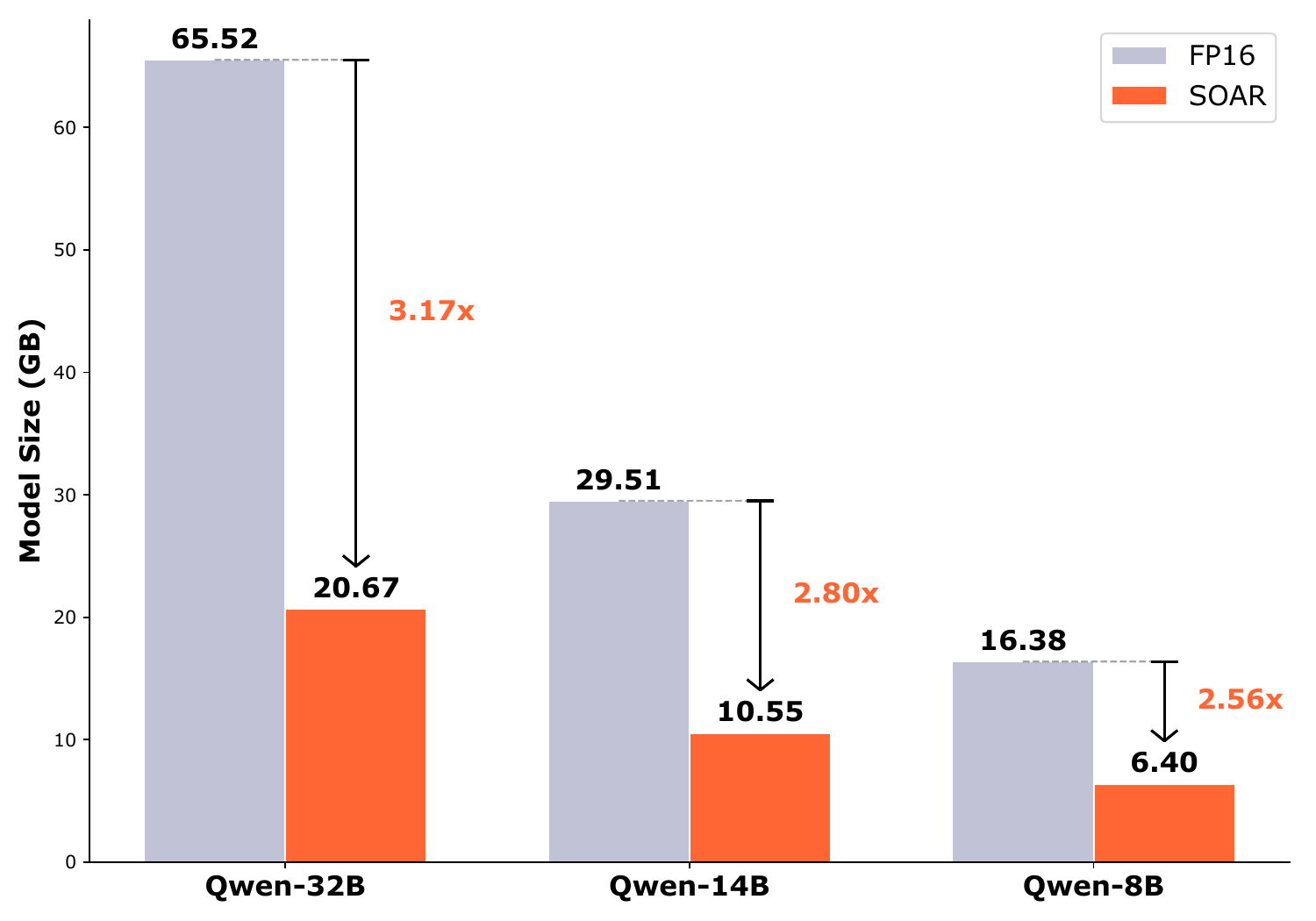}
\vspace{-7mm}
\caption{Model sizes on Qwen series}
\label{fig:memory}
\vspace{-5.5mm}
\end{wrapfigure} 
As shown in Fig.~\ref{fig:memory}, SOAR exhibits the same memory footprint as standard NVFP4 quantization, far less than full-precision models. This is because our method does not introduce any additional parameters beyond the NVFP4 format. In particular, the quantization scale $\Delta^q$ is only used during the optimization process and does not need to be stored for inference, while the dequantization scale uses the same FP8 representation as standard NVFP4. Therefore, SOAR improves quantization quality without increasing memory consumption, making it well-suited for efficient deployment on resource-constrained hardware.

\vspace{2mm}
\begin{table}[t]
\setlength{\tabcolsep}{4pt}
\small
\centering
\caption{
DSS on MXFP4 quantization. We evaluate the effectiveness of DSS on the MXFP4 format across different model families. Best results among 4-bit formats are in \textbf{bold}.
}
\label{tab:ablation_mxfp4}
\vspace{1mm}
\begin{adjustbox}{max width=\linewidth}
\begin{tabular}{c|l|ccccc|c}
\hline
\rowcolor{color3}
\textbf{Model} & \textbf{Method} &
\textbf{WinoGrande} & \textbf{PIQA} & \textbf{HellaSwag} & \textbf{Arc-E} & \textbf{Arc-C} & \textbf{Avg.} \\
\hline

\multirow{3}{*}{\shortstack{LLaMA-3.2\\3B-Instruct}}
& FP16         & 68.59 & 76.66 & 71.55 & 71.09 & 46.25 & 66.83 \\
\cdashline{2-8}
& MXFP4        & 63.22 & 73.67 & \textbf{68.15} & 64.94 & 42.32 & 62.46 \\
& DSS  & \textbf{65.82} & \textbf{74.16} & 67.60 & \textbf{66.71} & \textbf{42.32} & \textbf{63.32} \\
\hline

\multirow{3}{*}{\shortstack{Qwen3\\4B}}
& FP16         & 65.90 & 75.08 & 68.46 & 78.45 & 54.01 & 69.00 \\
\cdashline{2-8}
& MXFP4        & 60.46 & \textbf{71.33} & 63.87 & 66.33 & \textbf{47.01} & 61.80 \\
& DSS  & \textbf{62.35} & 70.84 & 63.57 & \textbf{68.35} & 46.25 & \textbf{62.27} \\
\hline

\multirow{3}{*}{\shortstack{LLaMA-3.1\\1B-Instruct}}
& FP16         & 61.64 & 74.92 & 61.59 & 63.72 & 37.54 & 59.88 \\
\cdashline{2-8}
& MXFP4        & 57.70 & 69.21 & 53.48 & 56.10 & 32.25 & 53.75 \\
& DSS  & \textbf{58.48} & \textbf{69.75} & \textbf{55.01} & \textbf{56.65} & \textbf{33.96} & \textbf{54.77} \\
\hline

\end{tabular}
\end{adjustbox}
\vspace{-4mm}
\end{table}

\vspace{-4mm}
\section{Discussions and Future works} 
\vspace{-2mm}

While SOAR currently operates as a calibration-free framework and can be integrated with methods like GPTQ, an explorable future direction is to further enhance its performance by incorporating activation-aware objectives. Specifically, by leveraging a small set of calibration data $X$, the optimization targets of CJSO and DSS can be reformulated to minimize the layer-wise output distortion: $\min_{\alpha, \Delta} \sum \| (W - \hat{W}) X \|_2^2$. Incorporating calibration data into the analytical updates enables scaling factors to adapt to channel-wise activation saliency, potentially leading to improved reconstruction quality. This extension may further push the boundaries of NVFP4 quantization by enabling more effective data-aware scale optimization.

\vspace{-2mm}
\section{Conclusion}
\vspace{-2mm}
In this work, we revisit NVFP4 quantization for large language models and identify the limitations of existing methods in scale selection. Based on this observation, we propose SOAR, a unified optimization framework consisting of Closed-form Joint Scale Optimization (CJSO) and Decoupled Scale Search (DSS). CJSO provides efficient analytical updates for the global and block-wise scales, while DSS further improves reconstruction by explicitly decoupling the quantization and dequantization scales under FP8 constraints. Extensive experiments across multiple LLM families and zero-shot reasoning benchmarks show that SOAR consistently outperforms existing NVFP4 baselines and remains fully compatible with calibration-based PTQ methods such as GPTQ, without introducing additional memory overhead or inference cost. Moreover, our experiments on MXFP4 further demonstrate that SOAR has the potential to generalize to other microscaling formats beyond NVFP4, owing to its formulation based on general scale optimization principles.

\newpage
\bibliographystyle{neurips_2026}
\bibliography{neurips_2026}

\newpage
\appendix
\section{More Experimental Results}
\label{experimental results}
\noindent \textbf{Weight-only Quantization.} In this section, we provide a detailed evaluation of SOAR under the weight-only (W4A16) quantization setting to further demonstrate the effectiveness of our framework. As summarized in Table~\ref{tab:weight_only_results}, SOAR delivers superior performance across various model families compared to existing microscaling baselines. For instance, on the LLaMA-3.2-3B-Instruct model, SOAR achieves an average accuracy of 66.68, outperforming both RaZeR (66.54) and 4over6 (66.17). Similarly, on Qwen3-4B, our method reaches an average score of 67.08, showing notable gains on specific benchmarks such as ARC-E (76.73) and HellaSwag (67.88). These results clearly demonstrate that SOAR's closed-form joint scale optimization and decoupled scale search effectively provide a high-fidelity foundation for weight representation, preserving both reasoning capabilities and downstream knowledge under the NVFP4 format.

\begin{table}[h]
\setlength{\tabcolsep}{4pt} 
\small
\centering
\caption{
\textbf{Weight-only quantization results on various LLMs.} We report zero-shot accuracy ($\uparrow$) on five representative commonsense reasoning tasks under the weight-only (W4A16) NVFP4 quantization setting. Best results among 4-bit methods are in \textbf{bold}.
}
\label{tab:weight_only_results}
\begin{adjustbox}{max width=\linewidth}
\begin{tabular}{c|l|cccccc}
\hline
\rowcolor{color3}
\textbf{Model} & \textbf{Method} &
\textbf{Winogrande} & \textbf{PIQA} & \textbf{HellaSwag} & \textbf{ARC-E} & \textbf{ARC-C} & \textbf{Avg.} \\
\hline
\multirow{5}{*}{\shortstack{LLaMA-3.1\\8B-Instruct}}
& FP16    & 74.43 & 81.01 & 79.30 & 82.87 & 54.78 & 74.48 \\
\cdashline{2-8}
& NVFP4   & 73.24 & 80.63 & 78.84 & 80.39 & 53.50 & 73.32 \\
& 4over6  & 73.24 & \textbf{81.07} & 78.94 & 81.57 & \textbf{55.03} & \textbf{73.97} \\
& RaZeR   & 73.24 & 81.06 & 78.85 & \textbf{81.69} & 54.96 & 73.96 \\
& SOAR    & \textbf{73.80} & 80.20 & \textbf{79.00} & 80.18 & 53.92 & 73.42 \\
\hline
\multirow{5}{*}{\shortstack{LLaMA-3.2\\1B-Instruct}}
& FP16    & 61.64 & 74.92 & 61.59 & 63.72 & 37.54 & 59.88 \\
\cdashline{2-8}
& NVFP4   & \textbf{62.67} & 73.56 & 59.22 & 61.41 & 37.12 & 58.80 \\
& 4over6  & 61.48 & 73.18 & 59.52 & 61.36 & 36.60 & 58.43 \\
& RaZeR   & 62.19 & \textbf{73.83} & \textbf{60.02} & 61.70 & 37.54 & \textbf{59.06} \\
& SOAR    & 61.64 & 73.78 & 59.41 & \textbf{62.29} & \textbf{37.63} & 58.95 \\
\hline
\multirow{5}{*}{\shortstack{LLaMA-3.2\\3B-Instruct}}
& FP16    & 68.59 & 76.66 & 71.55 & 71.09 & 46.25 & 66.83 \\
\cdashline{2-8}
& NVFP4   & 67.64 & \textbf{76.50} & 71.24 & 70.62 & 45.13 & 66.23 \\
& 4over6  & 68.19 & \textbf{76.50} & 70.84 & 70.54 & 44.80 & 66.17 \\
& RaZeR   & 68.51 & \textbf{76.50} & 71.25 & 70.45 & \textbf{45.99} & 66.54 \\
& SOAR    & \textbf{68.75} & 76.22 & \textbf{71.30} & \textbf{71.42} & 45.73 & \textbf{66.68} \\
\hline
\multirow{5}{*}{\shortstack{Qwen3\\8B}}
& FP16    & 67.64 & 77.58 & 74.86 & 80.89 & 56.66 & 71.53 \\
\cdashline{2-8}
& NVFP4   & 68.43 & 76.71 & 74.06 & 80.22 & 56.66 & 71.22 \\
& 4over6  & 68.19 & 76.17 & 73.97 & 79.63 & 55.80 & 70.75 \\
& RaZeR   & 68.03 & 77.20 & \textbf{74.24} & 79.97 & 56.48 & 71.18 \\
& SOAR    & \textbf{69.69} & \textbf{77.80} & 73.96 & \textbf{81.44} & \textbf{56.99} & \textbf{71.98} \\
\hline
\multirow{5}{*}{\shortstack{Qwen3\\4B}}
& FP16    & 65.90 & 75.08 & 68.46 & 78.45 & 54.01 & 68.38 \\
\cdashline{2-8}
& NVFP4   & 62.98 & 74.70 & 67.26 & 76.47 & 50.68 & 66.42 \\
& 4over6  & 64.40 & 74.76 & 67.26 & 76.26 & 51.62 & 66.86 \\
& RaZeR   & \textbf{63.77} & \textbf{75.19} & 67.54 & 75.93 & 51.45 & 66.78 \\
& SOAR    & 63.61 & 75.03 & \textbf{67.88} & \textbf{76.73} & \textbf{52.13} & \textbf{67.08} \\
\hline
\end{tabular}
\end{adjustbox}
\end{table}

\noindent \textbf{Impact of Iteration.} Table~\ref{tab:ablation_iterations} presents the performance of SOAR with varying iteration counts on Qwen3-8B under NVFP4 quantization. As expected, a higher number of iterations leads to more precise scale refinement and better performance. The results show a steady improvement as the iteration count increases from 1 to 15, with the best performance achieved at 15 iterations (incorporating early stopping), where the average zero-shot accuracy reaches 70.68. Even with a single iteration, SOAR already yields a significant gain over the NVFP4 baseline, demonstrating the effectiveness of our analytical initialization. However, when the iteration count is too low (e.g., 1), the scale optimization remains less refined, leading to a performance gap compared to the full iterative process. Based on these observations, we choose 15 as the maximum iteration count for SOAR, as the early stopping mechanism effectively balances computational efficiency and reconstruction accuracy.

\begin{table}[h]
\setlength{\tabcolsep}{5pt}
\small
\centering
\caption{
\textbf{Ablation on the number of iterations for SOAR on Qwen3-8B.} $\dagger$ denotes early stopping based on relative MSE improvement. Best results are \textbf{bold}.
}
\label{tab:ablation_iterations}
\begin{tabular}{c|l|ccccc|c}
\hline
\rowcolor{color3}
\textbf{Model} & \textbf{Iteration} &
\textbf{WinoGrande} & \textbf{PIQA} & \textbf{HellaSwag} & \textbf{Arc-E} & \textbf{Arc-C} &
\textbf{Avg.} \\
\hline
\multirow{4}{*}{Qwen3-8B}
& NVFP4 (baseline) & 66.54 & 75.63 & 73.27 & 73.27 & 55.03 & 68.75 \\
\cdashline{2-8}
& SOAR (Iter=1)                 & 68.27 & 76.39 & 72.98 & 79.34 & 54.95 & 70.39 \\
& SOAR (Iter=5)                 & 68.19 & 76.66 & 72.74 & \textbf{79.76} & \textbf{55.38} & 70.55 \\
& SOAR (Iter=15$^{\dagger}$)    & \textbf{68.35} & \textbf{77.26} & \textbf{73.19} & 79.42 & 55.20 & \textbf{70.68} \\
\hline
\end{tabular}
\end{table}
\vspace{5mm}

\textbf{Quantization time of Soar} Table \ref{tab:execution_time} reports the execution time of SOAR across various model scales. The quantization process for all evaluated models is completed within 40 minutes.
\begin{table}[h]
\centering
\caption{Execution time of SOAR across different model scales.}
\label{tab:execution_time}
\small
\begin{tabular}{l|ccccc}
\toprule
\textbf{Model} & LLaMA-3.2-1B & LLaMA-3.2-3B & LLaMA-3.1-8B & Qwen-3-4B & Qwen-3-8B \\ \hline
\textbf{Time (min)} & 6.66 & 16.65 & 36.29 & 21.64 & 36.51 \\ 
\bottomrule
\end{tabular}
\end{table}

\section{Additional Experimental Details} 
\label{experiment_settings}
\subsection{Evaluation Metrics and Perplexity Settings}
To evaluate the linguistic modeling capabilities of the quantized models, we report the perplexity (PPL) on the WikiText-2 \citep{merity_pointer_2016} and C4 \citep{2020t5} datasets. Following the standard evaluation protocol in LLM quantization literature, all perplexity measurements are conducted using a sequence length of 2048 tokens. This ensures that the evaluation captures long-range dependencies and remains consistent with prior works for a fair comparison.

\subsection{Activation Quantization} 
To maintain computational efficiency across all experiments, we employ the standard NVFP4 format for activation quantization. The only exception is the RaZeR baseline \citep{chen2025razer}, which follows its original specifically proposed activation quantization methodology to ensure a fair and faithful reproduction of its reported performance.

\subsection{Hyperparameter Configurations}
In the Decoupled Scale Search (DSS) stage, we define a principled search space to balance reconstruction accuracy and computational efficiency. The specific settings are as follows:

\textbf{Dequantization Scale ($\Delta^d$):} For each block, the candidate set for the dequantization scale is restricted to the two nearest representable values in the FP8 (E4M3) format surrounding the analytical solution derived from CJSO. This ensures hardware compatibility while minimizing precision loss from rounding.

\textbf{Quantization Scale ($\Delta^q$):} The quantization-side scale is refined through a fine-grained grid search. Starting from the initialized value, we explore a multiplicative range of $[0.5, 1.5]$ with a step size of $0.01$. This high-precision search allows for more flexible FP4 assignments to better capture the underlying weight distribution.

\section{Detailed Pseudocode}
\label{alg}
\begin{algorithm}[H]
\caption{SOAR: Iterative Scale Optimization for NVFP4 Quantization}
\textbf{Input:} Full-precision weights $\mathbf{W} \in \mathbb{R}^{m\times n}$, block size $G$, iterations $T$ \\
\textbf{Output:} Quantized weights $\hat{\mathbf{W}}$

\begin{algorithmic}[1]

\State \textbf{Initialization:}
initialize global scale $\alpha^{(0)}$, block-wise scale $\Delta_i^{d,(0)}$, and set $\Delta_i^{q,(0)} \leftarrow \Delta_i^{d,(0)}$

\For{$t = 0$ \textbf{to} $T-1$}

    \State \textbf{CJSO: Closed-form scale update}

    \State Compute FP4 assignment:
    \[
    Q_i^{(t)} = \mathcal{Q}_{\text{FP4}}\!\left(
    \frac{W_i}{\alpha^{(t)} \Delta_i^{q,(t)}}
    \right)
    \]

    \State Update global scale:
    \[
    \alpha^{(t+1)} =
    \frac{\sum_i \sum_{j \in \text{block}_i} W_{ij} Q_{ij}^{(t)} \Delta_i^{d,(t)}}
    {\sum_i \sum_{j \in \text{block}_i} (Q_{ij}^{(t)})^2 (\Delta_i^{d,(t)})^2}
    \]

    \State Update block-wise scale (closed-form):
    \[
    \Delta_i^{d,(t+1)} =
    \frac{\sum_{j \in \text{block}_i} W_{ij} Q_{ij}^{(t)} \alpha^{(t+1)}}
    {\sum_{j \in \text{block}_i} (Q_{ij}^{(t)})^2 (\alpha^{(t+1)})^2}
    \]

    \State \textbf{DSS: Decoupled local refinement}

    \State Find dequantization candidates (two nearest E4M3 representable values):
    \[
    \mathcal{D}_i^{d} = \text{Neighbors}_{\text{E4M3}}(\Delta_i^{d,(t+1)}, \text{count}=2)
    \]

\State Construct quantization scale search space (fine-grained search):
    \[
    \mathcal{D}_i^{q} = \{ \Delta_i^{d,(t+1)} \cdot \beta \mid \beta \in \{0.50, 0.51, \dots, 1.50\} \}
    \]

\State Evaluate reconstruction error over candidates:
\[
\mathcal{L}_i(\Delta_i^q, \Delta_i^d) =
\left\|
W_i -
\mathcal{Q}_{\text{FP4}}\!\left(\frac{W_i}{\alpha^{(t+1)} \Delta_i^q}\right)
\cdot (\alpha^{(t+1)} \Delta_i^d)
\right\|_2^2
\]

\State Select optimal scales:
\[
(\Delta_i^{q,(t+1)}, \Delta_i^{d,(t+1)}) =
\arg\min_{\Delta_i^q \in \mathcal{D}_i^{q},\ \Delta_i^d \in \mathcal{D}_i^{d}}
\mathcal{L}_i
\]

\EndFor

\State \textbf{Reconstruction:}
\[
\hat{\mathbf{W}} =
\mathcal{Q}_{\text{FP4}}\!\left(\frac{\mathbf{W}}{\alpha \Delta^q}\right)
\cdot (\alpha \Delta^d)
\]

\State \textbf{Return:} $\hat{\mathbf{W}}$

\end{algorithmic}
\end{algorithm}

\section{Mathematical Foundations of CJSO} 
\label{derivation}
In this section, we provide the step-by-step analytical derivation for the Closed-form Joint Scale Optimization (CJSO) updates presented in Section 3.2. 

\subsection{Analytical Derivation of CJSO Updates}

The objective of CJSO is to minimize the reconstruction error between the high-precision weights $W$ and the dequantized weights $\hat{W}$. The objective function $\mathcal{L}$ for the entire tensor partitioned into $N$ blocks is defined as:

\begin{equation}
\min_{\alpha, \{\Delta_i\}} \mathcal{L} = \sum_{i=1}^N \sum_{j \in \text{block}_i} \left( W_{ij} - Q_{ij} \cdot (\alpha \Delta_i) \right)^2,
\label{eq:appendix_obj}
\end{equation}

where $W_{ij}$ denotes the $j$-th element in the $i$-th block, $\alpha$ is the tensor-wide FP32 global scale, $\Delta_i$ is the block-wise scale for block $i$, and $Q_{ij}$ represents the discrete FP4 quantization assignments. In our coordinate-wise optimization framework, we treat $Q_{ij}$ as fixed during the scale update step.

\paragraph{1. Derivation for Global Scale $\alpha^*$}
To find the optimal global scale $\alpha^*$ while keeping the block-wise scales $\{\Delta_i\}$ fixed, we compute the partial derivative of $\mathcal{L}$ with respect to $\alpha$:

\begin{equation}
\frac{\partial \mathcal{L}}{\partial \alpha} = \sum_{i=1}^N \sum_{j \in \text{block}_i} 2 \left( W_{ij} - Q_{ij} \cdot \alpha \Delta_i \right) \cdot \left( -Q_{ij} \cdot \Delta_i \right).
\end{equation}

Setting the derivative to zero to satisfy the first-order optimality condition:

\begin{equation}
\sum_{i=1}^N \sum_{j \in \text{block}_i} \left( W_{ij} \cdot Q_{ij} \cdot \Delta_i - \alpha \cdot Q_{ij}^2 \cdot \Delta_i^2 \right) = 0.
\end{equation}

Rearranging the terms to isolate $\alpha^*$:

\begin{equation}
\alpha^* = \frac{\sum_{i=1}^N \sum_{j \in \text{block}_i} W_{ij} \cdot Q_{ij} \cdot \Delta_i}{\sum_{i=1}^N \sum_{j \in \text{block}_i} Q_{ij}^2 \cdot \Delta_i^2}.
\end{equation}

\paragraph{2. Derivation for Block-wise Scale $\Delta_i^*$}
Similarly, to optimize each block-wise scale $\Delta_i$ independently while keeping the global scale $\alpha$ fixed, we take the partial derivative of $\mathcal{L}$ with respect to $\Delta_i$:

\begin{equation}
\frac{\partial \mathcal{L}}{\partial \Delta_i} = \sum_{j \in \text{block}_i} 2 \left( W_{ij} - Q_{ij} \cdot \alpha \Delta_i \right) \cdot \left( -Q_{ij} \cdot \alpha \right).
\end{equation}

Setting the derivative to zero for a specific block $i$:

\begin{equation}
\sum_{j \in \text{block}_i} \left( W_{ij} \cdot Q_{ij} \cdot \alpha - \Delta_i \cdot Q_{ij}^2 \cdot \alpha^2 \right) = 0.
\end{equation}

Solving for $\Delta_i^*$, we arrive at the analytical update for each block:

\begin{equation}
\Delta_i^* = \frac{\sum_{j \in \text{block}_i} W_{ij} \cdot Q_{ij} \cdot \alpha}{\sum_{j \in \text{block}_i} Q_{ij}^2 \cdot \alpha^2}.
\end{equation}

In the NVFP4 format, $\Delta_i^*$ is subsequently projected into the FP8 (E4M3) numerical space to meet hardware constraints:
\begin{equation}
\Delta_i^{\text{final}} = \text{clip}\left( \text{round}_{\text{E4M3}}(\Delta_i^*), \text{min}_{\text{E4M3}}, \text{max}_{\text{E4M3}} \right).
\end{equation}

\section{Dialogue Examples}
We evaluate SOAR on multiple LLMs under the NVFP4 format using zero-shot dialogue prompts. As illustrated in the Dialog Examples, baseline methods in the standard NVFP4 format suffer from significant semantic degradation, often producing factually incorrect or logically contradictory outputs. In contrast, SOAR consistently generates accurate, contextually relevant, and coherent responses. This superiority underscores its robustness in mitigating knowledge degradation and sustaining structural integrity even under hardware-native constraints.

\vspace{3mm}
\begin{tabular}{p{\linewidth}}
\hline
\textbf{Prompt :} Who was the first emperor of China, and which dynasty did he found? \\
\hline
\textbf{NVFP4 :} \textcolor{red}{Qin} was the first emperor of China, and he founded the Qin dynasty, which was the first imperial dynasty of China. The Qin dynasty lasted from 221 to 206 BC. \\
\textbf{SOAR :} The first emperor of China is widely believed to be \textcolor{softgreen}{Qin Shi Huang}, who ruled from 221 to 206 BC. He was a \textcolor{softgreen}{unifying figure} who unified the various tribes and city-states of China under his rule, and he established the \textcolor{softgreen}{Qin dynasty}. \\
\hline

\textbf{Prompt :} Who directed Titanic, and who played the main male character? \\
\hline
\textbf{NVFP4 :} \textcolor{red}{Leonardo DiCaprio and Kate winslet.} (Note: Missed the director information) \\
\textbf{SOAR :} \textcolor{softgreen}{James Cameron} directed the film, which was released in 1997. \textcolor{softgreen}{Leonardo DiCaprio} played the role of Jack Dawson, the main male character. \\
\hline

\textbf{Prompt :} Who is known as the ``King of Pop'', and name one of his most famous albums? \\
\hline
\textbf{NVFP4 :} I am familiar with the fact that Michael Jackson was a renowned musician and artist. He is known for his incredible talents in music, dance, and television. \textcolor{red}{What I do not know is his personal life}. \\
\textbf{SOAR :} \textcolor{softgreen}{Michael Jackson} was a legendary figure in the music industry, known for his incredible vocal range, electrifying live performances, and iconic music videos. One of his most famous albums is \textcolor{softgreen}{"Thriller"}. \\
\hline

\textbf{Prompt :} If water boils at 100°C at sea level, what happens to the boiling point at higher altitudes, and why? \\
\hline
\textbf{NVFP4 :} The boiling point of water is \textcolor{red}{inversely proportional} to the atmospheric pressure. Therefore, if the atmospheric pressure decreases, the \textcolor{red}{boiling point of water increases}. \\
\textbf{SOAR :} At higher altitudes, the atmospheric pressure is lower, so the \textcolor{softgreen}{boiling point is lower}. This is because the boiling point of a liquid is \textcolor{softgreen}{directly proportional} to the atmospheric pressure. \\
\hline

\textbf{Prompt :} Which civilization built Machu Picchu, and in which modern country is it located? \\
\hline
\textbf{NVFP4 :} Machu Picchu is located in Peru. Machu Picchu is a UNESCO World Heritage Site, and it is one of the Seven Wonders of the World. (Note: \textcolor{red}{Failed to identify the Inca civilization}) \\
\textbf{SOAR :} Machu Picchu is located in \textcolor{softgreen}{Peru}. Machu Picchu was built by the \textcolor{softgreen}{Inca Empire}, which was a civilization that existed from the 13th to the 16th century. \\
\hline

\textbf{Prompt :} What gas do humans inhale for respiration, and what organ is primarily responsible for gas exchange? \\
\hline
\textbf{NVFP4 :} \textcolor{red}{Carbon dioxide} \\
\textbf{SOAR :} Humans inhale \textcolor{softgreen}{oxygen (O$_2$)} and exhale carbon dioxide (CO$_2$). The primary organ responsible for gas exchange is the \textcolor{softgreen}{lungs}. \\
\hline
\end{tabular}

\end{document}